# Driver Behavior Modelling at the Urban Intersection via Canonical Correlation Analysis

Zirui Li
School of Mechanical Engineering,
Beijing Institute of Technology
Beijing, China
3120195255@bit.edu.cn

Chao Lu
School of Mechanical Engineering,
Beijing Institute of Technology
Beijing, China
chaolu@bit.edu.cn

Cheng Gong
School of Mechanical Engineering,
Beijing Institute of Technology
Beijing, China
1120163553@bit.edu.cn

Jianwei Gong
School of Mechanical Engineering,
Beijing Institute of Technology
Beijing, China
gongjianwei@bit.edu.cn

Jinghang Li
School of Mechanical Engineering,
Beijing Institute of Technology
Beijing, China
3120180346@bit.edu.cn

Lianzhen Wei
School of Mechanical Engineering,
Beijing Institute of Technology
Beijing, China
1120160602@bit.edu.cn

*Abstract*—The urban intersection is a typically dynamic and complex scenario for intelligent vehicles, which exists a variety of driving behaviors and traffic participants. Accurately modelling the driver behavior at the intersection is essential for intelligent transportation systems (ITS). Previous researches mainly focus on using attention mechanism to model the degree of correlation. In this research, a canonical correlation analysis (CCA)-based framework is proposed. The value of canonical correlation is used for feature selection. Gaussian mixture model and Gaussian process regression are applied for driver behavior modelling. Two experiments using simulated and naturalistic driving data are designed for verification. Experimental results are consistent with the driver's judgment. Comparative studies show that the proposed framework can obtain a better performance.

## I. Introduction

Accurately modelling driver behaviors is important for intelligent transportation systems (ITS). To precisely predict the driver behavior, various machine learning methods have been developed based on deep learning [1], reinforcement learning [2, 3] and statistic learning [4-7]. Those methods are applied in car-following, lane-changing and other complicated scenarios.

The urban intersection is a typically dynamic and complicated traffic scenario, where exists pedestrians, vehicles and riders. Many methods are proposed to model behavior and interaction between participants, which are mainly divided into three categories: trajectories prediction, intelligent vehicle's planning and control and multi-agent interactions modelling.

In [8], the influence of different traffic participants are considered with a feed-forward artificial neural network called influence-network for trajectory prediction. [9] presents an interaction dataset for driver interactive trajectory study which collected trajectories in dynamic and complex scenarios worldwide. To estimate driver behavior in intersections, [10] proposed to model the driver behavior as a hybrid-state system, and hidden Markov model (HMM) is applied to estimate the driver behavior. In [11], a finite-state machine (FSM) is applied to decide the driving strategy for intersection scenarios. For advanced driver assistance systems (ADAS) in a centralized interaction control scenario, [12] proposed a personalized pace optimization algorithm for signalized intersections. Based on the game theory, [13] proposed a framework to model the interactive driver behavior at uncontrolled interactions, where both autonomous and human-driven vehicles are considered.

Although many kinds of research have been done in modelling driver behaviors at the intersection, a basic assumption of these research is that all traffic participants are related to driver behaviors of the host vehicle. In order to model the degree of correlation, the attention mechanism is applied in recent research [14, 15]. However, there is little work devoted to selecting useful features for modeling driver behavior by formulating correlation or influence. In this research, a canonical correlation analysis (CCA)-based framework is proposed, which selects features by analyzing the correlation between the host vehicle and surrounding traffic participants [16]. Gaussian mixture model(GMM) and Gaussian process regression(GPR) are used for driver behavior modelling [17-19]. The proposed framework is verified both in simulated environment and naturalistic driving dataset.

Main contributions of this research are two-fold:

1. A CCA-based framework is proposed for driver behavior modelling. The value of canonical correlation is applied for feature selection in driver behavior modelling process, which is formulated by GMR and GPR.

2. To verify the proposed framework, four scenarios at the intersection are designed in the simulated environment. And naturalistic driving dataset are collected from the urban intersection.

## II. Framework

The whole framework consists of three parts, including data collection, canonical correlation analysis, and driving behavior modeling. The process of data collection has two components: collect simulated driving data in CARLA and the on-road collection of naturalistic driving data. In the part of correlation analysis, the correlation coefficient of CCA is calculated for feature selection. In the third part, GPR and GMR are applied to predict driving behaviors of host vehicle. The problem formulation of CCA, GMR and GPR are presented in Section III. Meanwhile, the process of data collection is detailed in Section Ⅳ.

*This work was supported by the National Natural Science Foundation of China 61703041，SAIC Motor Industry-University-Research Collaboration Project 1813，and Jiangsu Key R&D Plan BE2018088.
Corresponding author: Jianwei Gong.

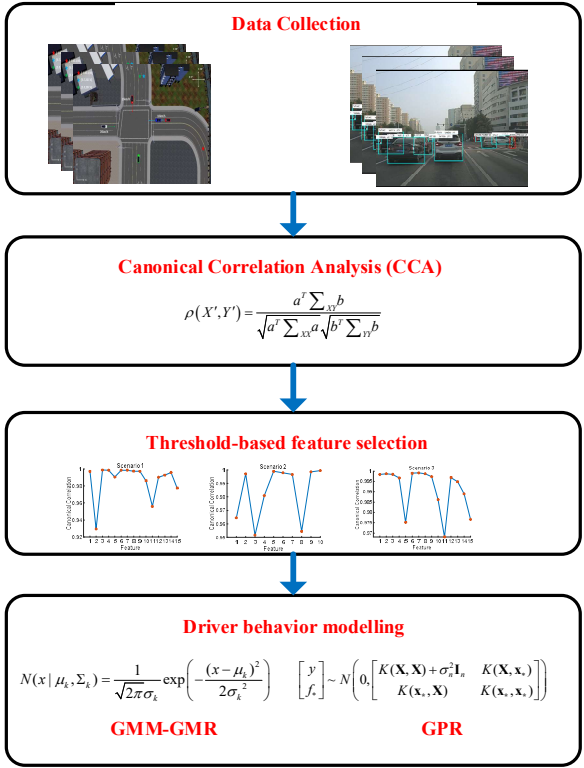

Fig.1. The overall illustration of proposed framework.

## III. METHODOLOGY

### A. Canonical Correlation Analysis (CCA)

CCA is a multivariate statistical analysis method that uses the correlation between comprehensive variable pairs to reflect the overall correlation between two sets of indicators. Given two column vectors with random variable, $\mathbf{X}=(x_1,\ldots,x_n)^\mathrm{T}$ and $\mathbf{Y}=(y_1,\ldots,y_m)^\mathrm{T}$, we can define mutual covariance matrix $\underset{\mathbf{XY}}{\Sigma}=cov(\mathbf{X},\mathbf{Y})$ of $n\times m$, where the element in position $(i,j)$ is covariance $cov(x_i,y_j)$. In CCA, a projection vector $a$ is applied to embedding $\mathbf{X}$ to the vector of one dimension. Similarly, the projection vector of $\mathbf{Y}$ is $b$. The vector obtained after the projection of $\mathbf{X}$ and $\mathbf{Y}$ is defined as $\mathbf{X'}$ and $\mathbf{Y'}$:

$$\mathbf{X'}=\mathbf{a}^\mathrm{T}\mathbf{X} \qquad (1)$$

$$\mathbf{Y'}=\mathbf{b}^\mathrm{T}\mathbf{Y} \qquad (2)$$

The optimization objective of CCA is to maximize $\rho(\mathbf{X'},\mathbf{Y'})$, that is, to find the vectors $a$ and $b$ when the maximum is reached. Assuming that the original data has been standardized, the optimization goal is formulated as:

$$\rho(\mathbf{X'},\mathbf{Y'})=\frac{cov(\mathbf{X'},\mathbf{Y'})}{\sqrt{D(\mathbf{X'})}\sqrt{D(\mathbf{Y'})}}=\frac{\mathbf{a}^\mathrm{T}\underset{\mathbf{XY}}{\Sigma}\mathbf{b}}{\sqrt{\mathbf{a}^\mathrm{T}\underset{\mathbf{XX}}{\Sigma}\mathbf{a}}\sqrt{\mathbf{b}^\mathrm{T}\underset{\mathbf{YY}}{\Sigma}\mathbf{b}}} \quad (3)$$

Since both numerator and denominator are increased by the same multiple, we can fix the denominator to 1, namely $\mathbf{a}^\mathrm{T}\underset{\mathbf{XX}}{\Sigma}\mathbf{a}=\mathbf{b}^\mathrm{T}\underset{\mathbf{YY}}{\Sigma}\mathbf{b}=1$, and the equation is transformed into:

$$\rho(\mathbf{X'},\mathbf{Y'})=\mathbf{a}^\mathrm{T}\underset{\mathbf{XY}}{\Sigma}\mathbf{b} \qquad (4)$$

The singular value decomposition (SVD) can be applied to solve this problem. Let $\mathbf{a}=(\underset{\mathbf{XX}}{\Sigma}\mathbf{u})^{-\frac{1}{2}}$, $\mathbf{b}=(\underset{\mathbf{YY}}{\Sigma}\mathbf{v})^{-\frac{1}{2}}$, Then the problem can be solved by applying $\mathbf{u}$ and $\mathbf{v}$ that correspond to the maximum singular value to $\mathbf{a}$, $\mathbf{b}$.

### B. Gaussian mixture model and Gaussian mixture regression (GMM-GMR)

The Gaussian mixture model (GMM) is a multivariate probability distribution model, which can estimate the probability density function of sample data. Given a group of sample data $\{x_1,x_2,x_3,\ldots,x_n\}$, it adopts the probability model with a weighted sum of multiple gaussian models, and can be described as:

$$p(x)=\sum_{k=1}^{K}p(k)p(x|k)=\sum_{k=1}^{K}\pi_k N(x|\mu_k,\Sigma_k) \qquad (5)$$

with

$$N(x|\mu_k,\Sigma_k)=\frac{1}{\sqrt{2\pi}\sigma_k}\exp\left(-\frac{(x-\mu_k)^2}{2\sigma_k^2}\right) \qquad (6)$$

where $\pi_k$ is the prior probability of the $k_{th}$ gaussian component and satisfies the constraint $\pi_k\geq 0, \sum_{k=1}^{K}\pi_k=1$. $N(x|\mu_k,\Sigma_k)$ is the single-gaussian distribution density function, in which $\Sigma_k$ is the standard deviation, $\sigma_k$ is the variance, and $\Sigma=\sigma^2$ is the standard deviation. To solve the probability density parameters, Expectation Maximization algorithm (EM) is widely used.

Based on GMM, GMR can be applied to represent continuous variable in the dataset. The output $\hat{h}$ can be estimated by maximizing the probability, which is shown as:

$$\hat{h}=\arg\max_{\boldsymbol{\theta}} M(x;\hat{\boldsymbol{\theta}}) \qquad (7)$$

where $\boldsymbol{\theta}=\{\pi_i,\mu_k,\Sigma_k\}_{k=1}^{K}$ is the GMM model, and $\hat{\boldsymbol{\theta}}$ is the output of GMM model. The probability of the output $\hat{h}$ Gaussian model is represented by:

$$\hat{\sigma}=\frac{\pi_k N_k(x,\mu_k,\Sigma_k)}{\sum_{k=1}^{K}\pi_i N_i(x,\mu_k,\Sigma_k)} \qquad (8)$$

### C. Gaussian Process Regression (GPR)

The characteristic of GP is determined by mean function $m(\mathbf{X}_{\mathrm{Env}})$ and covariance function $\mathrm{k}(\mathbf{X}_{\mathrm{Env}},\mathbf{X}'_{\mathrm{Env}})$:

$$m(\mathbf{X}_{\mathrm{Env}})=\mathrm{E}[f(\mathbf{X}_{\mathrm{Env}})] \qquad (9)$$

$$k(\mathbf{X}_{Env}, \mathbf{X}'_{Env}) = E[(f(\mathbf{X}_{Env}) - m(\mathbf{X}_{Env}))(f(\mathbf{X}'_{Env}) - m(\mathbf{X}'_{Env}))] \quad (10)$$

where $\mathbf{X}_{Env}, \mathbf{X}'_{Env} \in \mathbf{R}^d$ are arbitrary random variables. With the pre-process, $m(\mathbf{X}_{Env})$ is modified as zero mean function. Therefore, Gaussian process is formulated as follow:

$$f(\mathbf{X}_{Env}) \sim GP(m(\mathbf{X}_{Env}), k(\mathbf{X}_{Env}, \mathbf{X}'_{Env})) \quad (11)$$

As for the regression problem, considering the following equation with noise $\varepsilon$:

$$y = f(\mathbf{x}) + \varepsilon \quad (12)$$

where $\mathbf{x} \in \mathbf{R}^{N \times d}$ is input vector with $d$ dimensions and $\varepsilon$ is Gaussian noisy with $\varepsilon \in N(0, \sigma_n^2)$. The prior distribution of $y$ can be obtained and shown as follow:

$$\mathbf{Y}_{Op} \sim N(0, K(\mathbf{X}_{Env}, \mathbf{X}_{Env}) + \sigma_n^2 \mathbf{I}_n) \quad (13)$$

Combining Eq.(12) and Eq.(13), the joint distribution of prediction and observation is as follow:

$$\begin{bmatrix} \mathbf{Y}_{Op} \\ f_* \end{bmatrix} \sim N\left(0, \begin{bmatrix} K(\mathbf{X}_{Env}, \mathbf{X}_{Env}) + \sigma_n^2 \mathbf{I}_n & K(\mathbf{X}_{Env}, x^*_{Env}) \\ K(x^*_{Env}, \mathbf{X}_{Env}) & K(x^*_{Env}, x^*_{Env}) \end{bmatrix}\right) \quad (14)$$

where $K(\mathbf{X}_{Env}, \mathbf{X}_{Env}) = K_n = (k_{ij})$ is the symmetric positive definite covariance matrix and $K(\mathbf{X}_{Env}, \mathbf{X}_{Env}) \in \mathbf{R}^{n \times n}$. $k_{ij} = k(\mathbf{X}^i_{Env}, \mathbf{X}^j_{Env})$ is used to measure the correlation between $\mathbf{X}^i_{Env}$ and $\mathbf{X}^j_{Env}$. $K(\mathbf{X}_{Env}, x^*_{Env}) = K(x^*_{Env}, \mathbf{X}_{Env})^T$ is the covariance matrix between the test points $x_*$ and the input vector $x$. $K(x^*_{Env}, x^*_{Env})$ is the covariance matrix of $x_*$ itself and $\mathbf{I}_n \in n \times n$ is the unit matrix. The posterior distribution of GPR is as follow:

$$f_* | \mathbf{X}_{Env}, \mathbf{Y}_{Op}, x^*_{Env} \sim N(\bar{f}_*, cov(f_*)) \quad (15)$$

with

$$\bar{f}_* = K(x^*_{Env}, \mathbf{X}_{Env})[K(\mathbf{X}_{Env}, \mathbf{X}_{Env}) + \sigma_n^2 \mathbf{I}_n]^{-1} \mathbf{Y}_{Op} \quad (16)$$

$$cov(f_*) = K(x^*_{Env}, x^*_{Env}) - K(x^*_{Env}, \mathbf{X}_{Env}) \\ \times [K(\mathbf{X}_{Env}, \mathbf{X}_{Env}) + \sigma_n^2 \mathbf{I}_n]^{-1} K(\mathbf{X}_{Env}, x^*_{Env}) \quad (17)$$

where $\bar{f}_*$ and $cov(f_*)$ are mean and square deviation at test point $x_*$. Here we choose Gaussian kernel as the covariance function:

$$K(\mathbf{X}, \mathbf{X}') = \exp(-\frac{\|\mathbf{X} - \mathbf{X}'\|}{\sigma_f^2}) \quad (18)$$

where $\sigma_f$ is kernel width. For the model above, parameters $\boldsymbol{\theta} = \{\sigma_f, \sigma_n\}$ are undetermined which are essential for GPR in specific scenarios. Minimizing the negative log marginal likelihood function $L(\boldsymbol{\theta}) = -\log(p(y|\mathbf{X}, \boldsymbol{\theta}))$ is a common nethod to set hyper-parameters $\boldsymbol{\theta}$. The expression of $L(\boldsymbol{\theta})$ is:

$$L(\boldsymbol{\theta}) = \frac{1}{2} y^T \mathbf{C}^{-1} y + \frac{1}{2} \log |\mathbf{C}| + \frac{n}{2} \log 2\pi \quad (19)$$

And the partial derivatives of the marginal likelihood is

$$\frac{\partial L(\boldsymbol{\theta})}{\partial \theta_i} = \frac{1}{2} tr((\boldsymbol{\alpha}\boldsymbol{\alpha}^T - \mathbf{C}^{-1}) \frac{\partial \mathbf{C}}{\partial \theta_i}) \quad (20)$$

with $\mathbf{C} = \mathbf{K}_n + \sigma_n^2 \mathbf{I}_n$ and $\boldsymbol{\alpha} = (\mathbf{K} + \sigma_n^2 \mathbf{I}_n)^{-1} y = \mathbf{C}^{-1} y$.

## IV. EXPERIMENTS

### A. Experiment I: Simulated driving data

In order to verify the proposed framework in the driver behavior modelling. In Experiment I, CARLA platform is used for simulated data collection. As shown in Fig.2, four scenarios are designed in the simulation. And traffic participants include one host vehicle, three surrounding vehicles and one pedestrian.

Table I Statistics of features used for driver behavior modelling

| Items | Value |
|---|---|
| Traffic participants | Host vehicle, Surrounding vehicle 1,2 and 3 |
| Features | Longitudinal location, lateral location, heading angle, yaw velocity, longitudinal velocity, and longitudinal acceleration. |

The T-shaped and the cross-shaped are two common types of intersections in urban roads. In Experiment I, both intersections are designed for the simulated driving data collection. In order to reflect the complexity of intersections in the simulated environment. Three surrounding vehicles and one pedestrian are added in the intersections. The following features of participants are recorded for driver behavior modelling: positions, velocity and acceleration. Meanwhile, in the simulated scenario, velocities of surrounding vehicles and pedestrians are randomly selected.

In addition, the motion of pedestrians will also influence the decision-making process of drivers in the host vehicle, so it is necessary to design the speed and moving routes of pedestrians rationally.

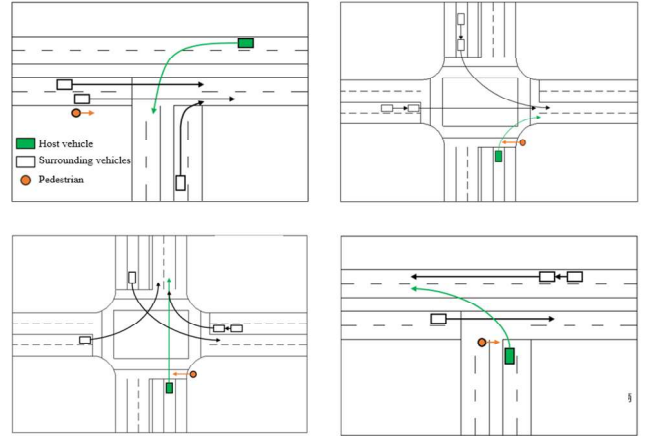

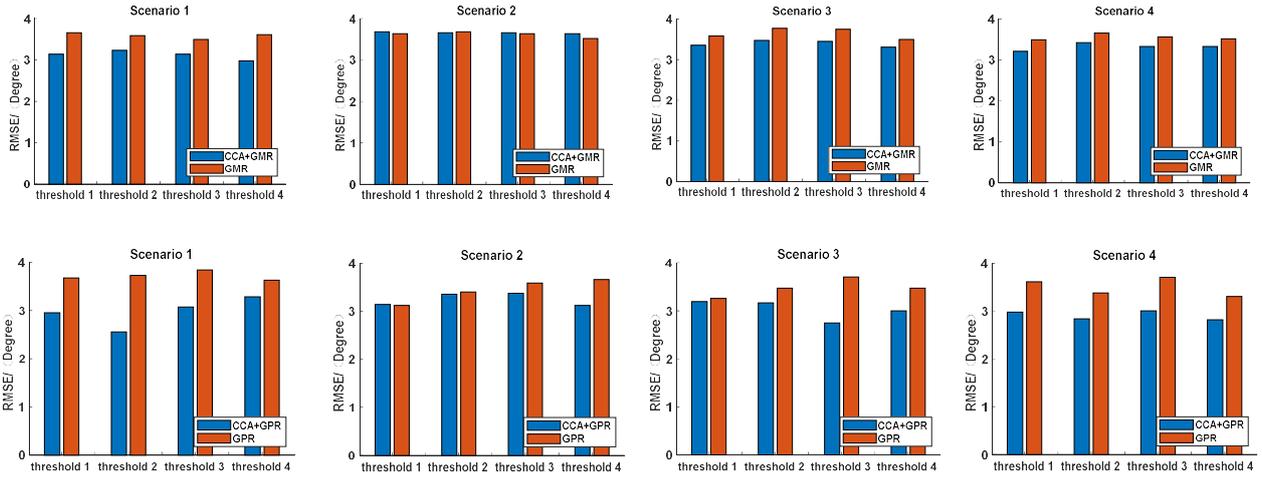
Fig.3. Experimental results for simulated driving data. (Experiment I)

Fig.2. The schematic diagram for four intersections in the simulated environment.

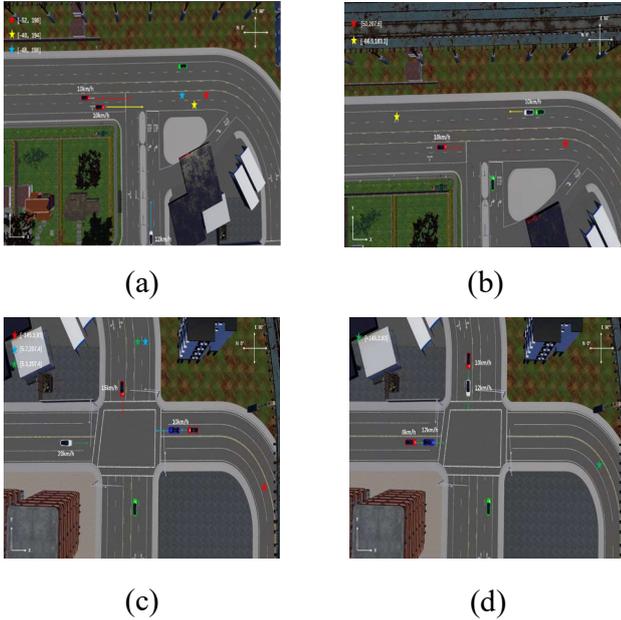

(a)          (b)

(c)          (d)

Fig.4. Four scenarios developed in CARLA.

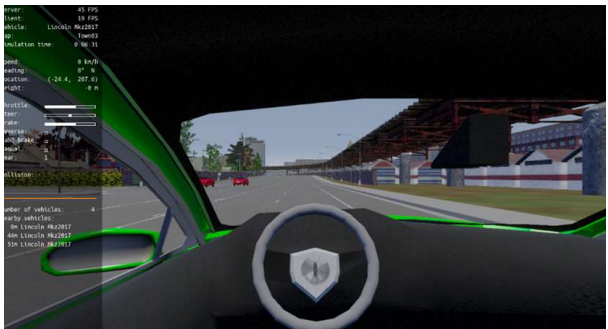

Fig.5. The view of human drivers in the data collection.

For the four traffic scenarios mentioned above, this study collects data through the Logitech G27 controller and Carla simulation environment. To ensure the validity of data and the consistency of driving data in distribution, all the collections are completed by a single driver. Each scene is repeatedly collected 50 times, which means that four scenes are collected 200 times in total.

Fig.3 illustrates results of Experiment I. Test scenarios 1 to 4 correspond to four intersection scenarios shown in Fig.2 and Fig.4, respectively. Four thresholds reflect the criteria of feature selection based on CCA, which are 0.80, 0.85, 0.90 and 0.95. The experimental results indicate that the proposed framework, CCA+GMR and CCA+GPR, can achieve better results in most cases. More specifically, in the second scenario, the above two frameworks do not significantly reduce the RMSE value but have obvious effect on other scenarios.

### B. Experiment Ⅱ: Naturalistic driving data

In order to collect naturalistic driving data for validation, the experimental vehicle used for data collection is shown in Fig.6. The sensor equipment is same as [20].

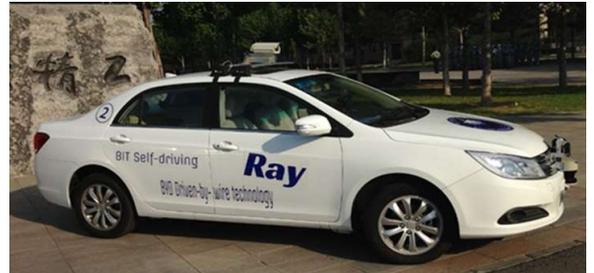

Fig.6. The experimental vehicle for on-road data collection.

Based on naturalistic driving data collected in the urban intersection, CCA is applied for the correlation analysis between possible influence factors and the behavior of host vehicle. Experimental results are presented in Fig.7 and Fig.8. The y-axis stands for the canonical correlation. The x-axis stands for different traffic participants.

In the above results, different target numbers represent the impact of different moving objects on the host vehicle in on-board test. The main influencing factors obtained in this section are basically consistent with the influence of moving targets on horizontal and vertical control in video playback, which can reflect that the typical correlation analysis method used in this section can relatively effectively express the influencing factors of urban intersections using test data.

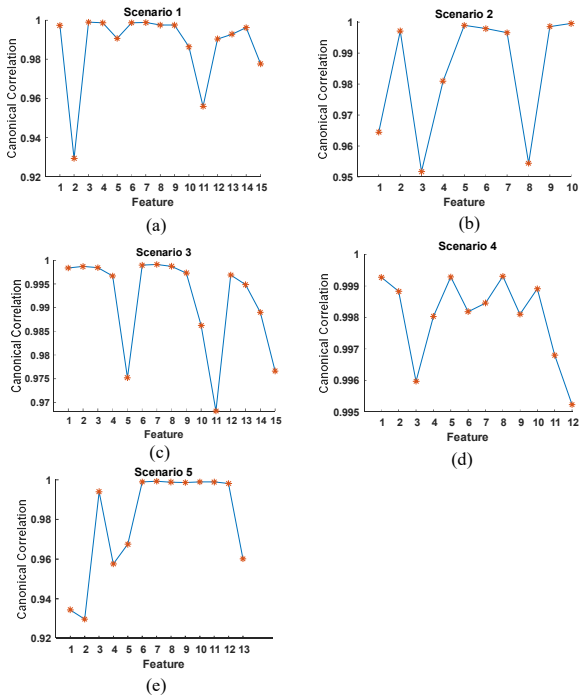

Fig.7. Experimental results for lateral driving behaviors. (Experiment Ⅱ)

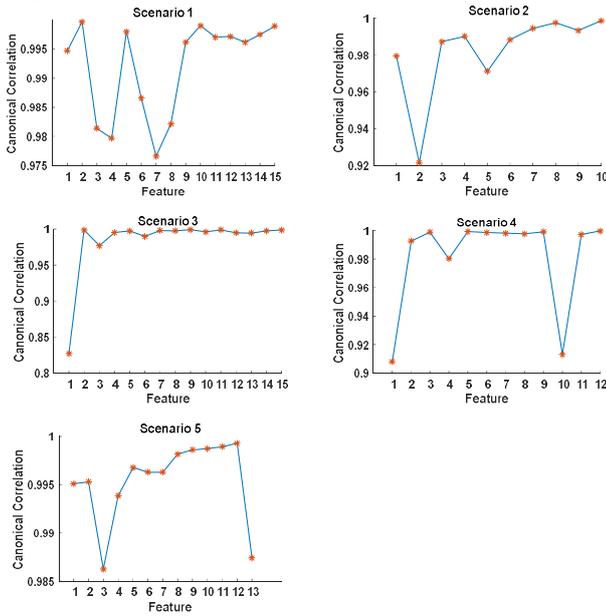

Fig.8. Experimental results for longitudinal driving behaviors. (Experiment Ⅱ)

## V. CONCLUSION

This research proposes a canonical correlation analysis (CCA)-based framework to model driver behaviors at the urban intersection. Based on the correlation analysis, the influence of alternative features is detailed, which is applied for feature selection and driver behavior modelling. Two experiments are designed and conducted by using simulated and naturalistic driving data. Comparative results with simulated data demonstrate that CCA-based framework obtains a better performance. And experimental results with naturalistic data is consistent with the driver's judgment. The main possible application is to systematically analysis influence factors in the process of driver behavior modelling.

Besides the application in the lane-changing scenario, as a general method, the proposed framework can also be applied to many other scenarios.